\documentclass[lettersize,journal]{IEEEtran}
\usepackage{amsmath,amsfonts}
\usepackage{algorithmic}
\usepackage{algorithm}
\usepackage{array}
\usepackage[caption=false,font=normalsize,labelfont=sf,textfont=sf]{subfig}
\usepackage{textcomp}
\usepackage{stfloats}
\usepackage{url}
\usepackage{verbatim}
\usepackage{graphicx}
\usepackage{cite}
\usepackage{enumitem}
\usepackage{xcolor}
\usepackage{tikz}
\usepackage{float}
\usepackage{booktabs}
\usepackage{multirow}
\usepackage[table,xcdraw]{xcolor}
\usepackage{pgfplots}
\pgfplotsset{compat=newest}
\pgfplotsset{plot coordinates/math parser=false}

\newtheorem{rem}{Remark}

\usepackage{hyperref}
\usepackage{lipsum}
\usepackage{bm}
\usepackage{siunitx}
\usepackage{threeparttable}

\def\BibTeX{{\rm B\kern-.05em{\sc i\kern-.025em b}\kern-.08em
    T\kern-.1667em\lower.7ex\hbox{E}\kern-.125emX}}

\begin{document}

\title{Real-Time Conformal-Seeded Hybrid Inverse Kinematics for Offset Redundant Manipulators}

\author{\IEEEauthorblockN{Duc Cuong Vu, \IEEEmembership{Student Member, IEEE}, Van Tung Nguyen, Duc Hai Nguyen,\\ Manh Cuong Nguyen, Vu Trung Tran, and Minh Nhat Vu, \IEEEmembership{Member, IEEE}.}

\thanks{Duc Cuong Vu is with VinRobotics, Hanoi 100000, Vietnam and the School of Electrical and Electronic Engineering, Nanyang Technological University 639798, Singapore (e-mail: {duccuong001@e.ntu.edu.sg}).}
\thanks{Van Tung Nguyen, Duc Hai Nguyen, Manh Cuong Nguyen, and Vu Trung Tran are with VinRobotics JSC, Hanoi 100000, Vietnam.}
\thanks{Minh Nhat Vu is with Center for AI Research (CAIR), VinUniversity, Hanoi 100000, Vietnam (e-mail: {minh.vn@vinuni.edu.vn}).}
\thanks{{Corresponding author: Minh Nhat Vu.}}
}

\markboth{IEEE Robotics and Automation Letters}%
{Vu \MakeLowercase{\textit{et al.}}: Real-Time Conformal-Seeded Hybrid Inverse Kinematics for Offset Redundant Manipulators}

\maketitle

\begin{abstract}
This paper presents a conformal-seeded hybrid strategy for solving inverse kinematics of {offset, redundant 7-DoF robot arms of the humanoid class}. Analytical inverse kinematics (AIK) provides closed-form solutions with very low computational cost. However, for offset kinematic structures, the exact closed-form solution is generally unavailable, and practical AIK must rely on an approximate or simplified kinematic model. In contrast, numerical inverse kinematics (NIK) can achieve high-precision solutions on the full kinematic model. However, its convergence is highly sensitive to initialization. To overcome these limitations, we propose a two-stage hybrid inverse kinematics framework with \emph{conformal-calibrated seed selection}. First, an approximate analytical model efficiently enumerates a finite set of candidate joint solutions. Second, we rank these candidates using a lightweight {\color{black}learned} predictor of post-refinement difficulty{\color{black}, wrapped by split-conformal prediction into a calibrated upper bound that serves as the selection score}. The best-ranked seed is then refined using a Levenberg-Marquardt solver on the full kinematic model. The proposed method combines fast candidate generation, {\color{black}learned seed ranking with a calibrated difficulty bound}, and accurate numerical refinement, achieving real-time performance of less than \SI{40}{\micro\second} and a success rate of \SI{100}{\percent} in our evaluation on reachable targets. We validate the approach through large-scale stochastic simulation across the workspace and experimental demonstrations with motion planning on a humanoid robot arm. Demonstration videos are available at https://youtu.be/aeiBmw1XRbw.
\end{abstract}

\begin{IEEEkeywords}
Redundant manipulator, Inverse kinematics, Analytical methods, Numerical optimization, Hybrid algorithms, Conformal-prediction
\end{IEEEkeywords}

\section{Introduction}

{Modern robotic systems increasingly require inverse kinematics (IK) solvers that are both computationally efficient and robust across diverse kinematic structures. Applications such as industrial manipulation}\cite{kaur2024trajectory}, {trajectory optimization} \cite{beck2024model}{, trajectory planning} \cite{gattringer2022time}{, require reliable real-time IK computation under redundancy, joint limits, and complex workspace constraints. These requirements become particularly challenging for robots with high degrees of freedom or nontrivial kinematic structures, where conventional IK methods may exhibit unstable convergence behavior or insufficient real-time reliability.} {This work targets offset 7-DoF redundant arms of the spherical-shoulder, offset-wrist class widely adopted in humanoid and dual-arm platforms, for which exact closed-form IK is unavailable and purely numerical solving is initialization-sensitive; the humanoid arm studied throughout serves as the representative instance.}

\textit{Analytical approach:} For robots with relatively simple kinematic structures \cite{zheng2024analytical}, certain classes of manipulators \cite{ostermeier2025automatic}, or well-structured redundant manipulators with favorable kinematic properties \cite{vu2023machine}, closed-form Analytical Inverse Kinematics (AIK) solutions can be derived. By exploiting geometric relationships and kinematic symmetries, {\emph{AIK provides exact solutions with extremely low computational cost and strong numerical stability}}, making them particularly attractive for real-time control and industrial applications. {However, \emph{AIK is limited}} \cite{aristidou2018inverse}{\emph{ when considering manipulators with high degrees of freedom, kinematic redundancy, or calibrated systems with kinematic parameter offsets}}. {Explicit analytical solutions are derived from an ideal kinematic chain model; practical factors such as mechanical tolerances, assembly errors, or calibration updates modify the robot description (DH parameters or URDF model), rendering the analytical solution inaccurate or even invalid.}

\begin{figure}
    \centering
    \scalebox{0.9}{\input{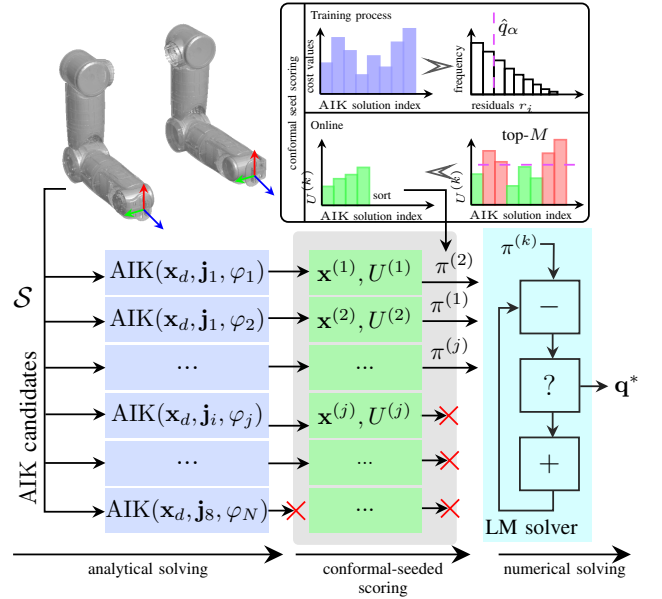}}
    \caption{Overview of the proposed novel \textbf{\emph{Conformal-Seeded Hybrid Inverse Kinematics (CS-HIK)}} framework. The method combines approximate analytical inverse kinematics with numerical refinement{\color{black}: each candidate receives a conformal upper bound $U^{(k)}$ on its predicted refinement difficulty and the candidates are refined in increasing order of $U^{(k)}$ up to the fail-safe budget $M$; no candidate is removed by a threshold}, thereby ensuring robust convergence and real-time performance across the entire workspace.}
    \label{fig:hook}
\end{figure}

\textit{Numerical approach}: To overcome these limitations, Numerical Inverse Kinematics (NIK) methods have been widely adopted 
\cite{zhang2025learning, Zakka_Mink_Python_inverse_2025, sundaralingam2023curobo},
in which the IK problem is typically formulated as a nonlinear optimization problem accounting for joint limits, motion smoothness, manipulability, etc. \cite{rakita2018relaxedik}. {\emph{A major advantage of numerical methods lies in their generality and flexibility}, which can operate with many types of robots} \cite{kim2025pyroki, beeson2015trac, bruyninckx2001open}, {requiring only the forward kinematics to formulate the optimization problem.} {Despite their flexibility, \emph{NIK methods suffer from several limitations, particularly regarding initial guess selection and computational time}}. {A poor initial guess may cause the solver to become trapped in local minima or near kinematic singularities}
\cite{nakamura1986inverse}. 
{In addition, iterative numerical solvers often exhibit unpredictable computational times on the order of milliseconds}\cite{aristidou2018inverse}.

{\textit{Hybrid approach:} Regarding hybrid approaches, the survey in} \cite{lakshmi2026decade} {shows that several studies have attempted to combine analytical and numerical IK methods. For example,} \cite{jin2020efficient} {employs Particle Swarm Optimization to estimate the elbow orientation; however, the execution time remains on the millisecond scale and the positioning accuracy is still limited. In} \cite{ananthanarayanan2013real, ananthanarayanan2015real}, {the authors analytically compute the wrist and shoulder joint configurations, followed by an iterative method to solve the elbow configuration. Although this approach achieves relatively fast execution, it still depends heavily on the initial guess, requires a spherical wrist structure, and exhibits large errors near workspace boundaries. Furthermore, in} \cite{kucuk2014inverse}, {the authors partially derive analytical solutions for the wrist joints while solving the remaining joints using the Newton--Raphson method. A common limitation of these approaches is that they mainly focus on reducing the dimensionality of the numerical optimization problem, while the initialization process for iterative refinement still relies on heuristic or random strategies. Consequently, the optimization process may still become trapped in local minima or require a large number of iterations for convergence.}

{In this study, motivated by the speed and robustness of analytical approaches, together with the generality and flexibility of iterative methods, we adopt a \emph{\textbf{H}ybrid \textbf{I}nverse \textbf{K}inematics (HIK)} paradigm that leverages the complementary strengths of both.} In a typical HIK framework, an approximate AIK {(quasi-analytical IK)} solution is first used to generate candidate initial seeds, which are {selected and} subsequently refined by an iterative numerical solver such as the Levenberg--Marquardt (LM) algorithm. {By leveraging analytical initialization,} HIK methods can significantly improve convergence robustness compared to purely numerical approaches.

{However, a key limitation of existing HIK methods is that joint configuration selection mechanisms (\emph{seed selection}) often rely on heuristic criteria, such as minimum joint displacement or maximum manipulability, which do not explicitly account for the convergence behavior of the numerical solver and may therefore lead to suboptimal performance.} {To address this challenge, this study proposes an additional seed-selection mechanism integrated into the HIK framework, called} \textit{\textbf{C}onformal-\textbf{S}eeded \textbf{H}ybrid \textbf{I}nverse \textbf{K}inematics (CS-HIK)}, that systematically identifies analytical seeds most compatible with the subsequent numerical refinement process {(shown in Fig. }\ref{fig:hook}{)}. The proposed approach leverages conformal prediction to evaluate the reliability of candidate solutions and select seeds that are more likely to converge successfully during numerical optimization. As a result, the method aims to maximize the convergence success rate while maintaining real-time computational performance across the entire robot workspace. {The framework addresses pose-only IK queries, the prevailing industrial setting in which the solver itself resolves the redundancy; the arm angle is nevertheless exposed as an explicit input of the analytical stage (Section~III-A), so a higher-level planner may command it, e.g.\ for obstacle avoidance or posture optimization.}

{The proposed quasi-AIK, executed within sub-$\mu\mathrm{s}$ time, requires only a few LM iterations for convergence, addressing the limitations of both AIK (\emph{offset, post-calibrated robots}) and NIK (\emph{initialization sensitivity and computational cost}).}

The main contributions of this paper are summarized as follows:
\begin{itemize}
        \item {A novel \textit{Conformal-Seeded Hybrid Inverse Kinematics (CS-HIK)} framework, whose core contribution is a {\color{black}learned seed-selection mechanism: a lightweight predictor of post-Levenberg--Marquardt refinement difficulty ranks the quasi-AIK candidate seeds, and split-conformal prediction wraps it into a calibrated, distribution-free upper bound that serves as the selection score. This score} identifies reliable initializations for LM refinement on the full kinematic model, yielding fast and robust IK performance with joint-limit awareness.}

    \item Extensive evaluation and benchmarking, including full-workspace Monte-Carlo analysis, comparisons with representative inverse kinematics methods {for an example humanoid robot arm}, and real-time experimental validation, demonstrating improved robustness, accuracy, and computational efficiency.
\end{itemize}

The remainder of this paper is organized as follows. Section~II presents the kinematic model and constraints of the redundant manipulator under consideration. Section~III introduces the proposed CS-HIK framework, including the derivation of the approximate analytical IK and the conformal seed selection strategy. Section~IV reports simulation, benchmarks, and experimental results. Finally, Section~V concludes the paper and discusses future research directions.
\begin{figure}
\centering
\includegraphics[width=0.9\linewidth]{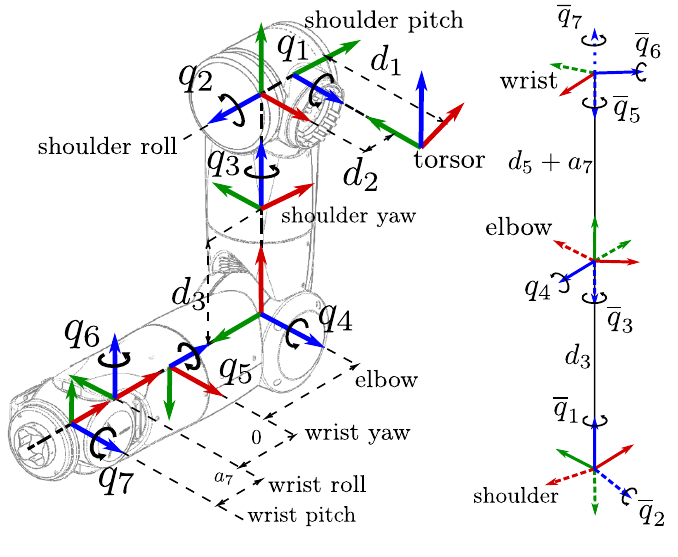}
\caption{Ideal kinematic configuration of the humanoid robot arm (right), and the simplified kinematics chain (left)}
\label{fig: robot configuration}
\end{figure}

\section{System Modeling}

This section introduces the kinematic structure of the robotic platform considered in this work. The manipulator is a seven-degree-of-freedom (7-DOF) serial robotic arm, enabling a wide range of end-effector positions and orientations within its workspace. The overall kinematic architecture of the manipulator is illustrated in Fig.~\ref{fig: robot configuration}, which shows the serial arrangement of links and revolute joints from the base to the end-effector. The corresponding modified Denavit--Hartenberg (MDH) parameters describing the nominal kinematic model are listed in Table~\ref{tab:dh_parameters_rad}.

\begin{table}
\centering
\caption{Ideal MDH parameters of the manipulator}
\renewcommand{\arraystretch}{1.3}
\begin{tabular}{ccccc}
\hline
\hline
\cellcolor[HTML]{EFEFEF}Link & \cellcolor[HTML]{EFEFEF}$\alpha_i$ (rad) & \cellcolor[HTML]{EFEFEF}$a_i$ (m) & \cellcolor[HTML]{EFEFEF}$\theta_i$ (rad) & \cellcolor[HTML]{EFEFEF}$d_i$ (m) \\
\hline
\hline
1 & ${5\pi}/{9}$ & 0 & $q_1 - {\pi}/{2}$ & -0.20624 \\
\cellcolor[HTML]{EFEFEF}2 & \cellcolor[HTML]{EFEFEF}${\pi}/{2}$ & \cellcolor[HTML]{EFEFEF}0 & \cellcolor[HTML]{EFEFEF}$q_2 + {5\pi}/{9}$ & \cellcolor[HTML]{EFEFEF}-0.009 \\
3 & $-{\pi}/{2}$ & 0 & $q_3 + {\pi}/{2}$ & -0.2247 \\
\cellcolor[HTML]{EFEFEF}4 & \cellcolor[HTML]{EFEFEF}${\pi}/{2}$ & \cellcolor[HTML]{EFEFEF}0 & \cellcolor[HTML]{EFEFEF}$q_4 + {\pi}/{2}$ & \cellcolor[HTML]{EFEFEF}0 \\
5 & ${\pi}/{2}$ & 0 & $q_5 + {\pi}/{2}$ & -0.2196 \\
\cellcolor[HTML]{EFEFEF}6 & \cellcolor[HTML]{EFEFEF}${\pi}/{2}$ & \cellcolor[HTML]{EFEFEF}0 & \cellcolor[HTML]{EFEFEF}$q_6 + {\pi}/{2}$ & \cellcolor[HTML]{EFEFEF}0 \\
7 & ${\pi}/{2}$ & -0.01 & $q_7$ & 0 \\
\hline
\hline
\end{tabular}%
\label{tab:dh_parameters_rad}
\end{table}

\begin{rem}
The kinematic structure and MDH parameters shown in Table~\ref{tab:dh_parameters_rad} represent the \textit{ideal robot model}. This kinematic configuration is widely adopted in many humanoid and dual-arm robotic systems \cite{zhong2023inverse}. \textit{In practice, after kinematic calibration, the actual geometric parameters of the robot may deviate from this ideal model due to manufacturing tolerances, assembly errors, and structural imperfections.} Consequently, the original MDH parameters may no longer accurately represent the true forward kinematics (FK) of the robot. Relying solely on the nominal MDH model for FK computation can therefore introduce modeling errors, which may degrade the accuracy of IK solutions derived from this idealized structure.
\end{rem}

Accordingly, the generalized coordinate vector of the robot arm is defined as
\begin{align}
\mathbf{q} =
\begin{bmatrix}
q_1 & q_2 & q_3 & q_4 & q_5 & q_6 & q_7 \label{eq:generalize coordiante}
\end{bmatrix}^\top
\in \mathcal{Q} \subset \mathbb{R}^{7},
\end{align}
where each element $q_i$ denotes the rotation angle of the $i$-th revolute joint, and the admissible joint space is given by
\begin{align}
\mathcal{Q} = \{ \mathbf{q} \in \mathbb{R}^{7} | \mathbf q_{\min} < \mathbf q < \mathbf q_{\max}\},\label{eq:set q}
\end{align}
{with $\mathbf{q}_{\min} = -(3.14, 2.36, 2.97, 0.52, 2.97, 0.35, 0.96)\,\mathrm{rad}$, and $\mathbf{q}_{\max} = (1.05, 0.0, 2.97, 1.57, 2.97, 0.35, 0.96)\,\mathrm{rad}$} for our manipulator.

\section{Hybrid Inverse Kinematics}
In this section, IK solution for the offset redundant manipulator is presented. First, the kinematic chain of the manipulator is simplified by reducing and reparameterizing several modified Denavit--Hartenberg (MDH) parameters, such as $a_7$ and the joint variables $q_i$, in order to obtain a tractable analytical formulation. Based on this simplified model, approximate AIK solutions are derived to generate candidate joint configurations. These candidate solutions are subsequently refined through a numerical optimization process, where the refined solutions are ranked and selected according to different criteria, including conventional heuristic-based strategies (e.g., joint displacement minimization and manipulability maximization) as well as the proposed conformal-seeded selection method.

\subsection{Analytical Inverse Kinematics (AIK)}

Closed-form analytical inverse kinematics solutions typically require robot structures with simple geometric properties, such as intersecting joint axes that enable kinematic decoupling. However, practical robotic manipulators often exhibit structural offsets introduced by mechanical design constraints and kinematic calibration, resulting in non-intersecting adjacent joint axes (see Fig.~\ref{fig: robot configuration}). These structural deviations make the derivation of exact closed-form analytical inverse kinematics solutions difficult or even infeasible.

To address this issue, a simplified kinematic model is adopted in this work. In this model, a bias configuration $\mathbf{q}_b$ is introduced, and the shoulder and wrist joint axes are assumed to intersect at common points. This approximation preserves the dominant geometric characteristics of the manipulator while enabling the derivation of a tractable analytical inverse kinematics formulation. Based on this simplified kinematic chain, the analytical inverse kinematics solutions can be expressed as \cite{vu2023machine}
\begin{align}
    \mathbf{q} = \mathrm{AIK}(\mathbf{x}_d,\mathbf{j},\varphi),
    \label{eq:aik}
\end{align}
where $\mathbf{x}_d$ denotes the desired end-effector pose, $\mathbf{j} \in \{-1,+1\}^3$ represents the discrete configuration parameters, and $\varphi \in [0,2\pi)$ is the null-space parameter associated with the redundant degree of freedom. {In practice, $\varphi$ is discretized uniformly with spacing $\Delta\varphi$, yielding $n_\varphi = \lceil 2\pi / \Delta\varphi \rceil$ swivel samples.} The function $\mathrm{AIK}(\mathbf{x}_d,\mathbf{j},\varphi)$ provides the corresponding joint configuration for a given redundancy configuration. {Since $\varphi$ enters \eqref{eq:aik} explicitly, the redundancy is exposed at the solver interface: a commanded arm angle $\varphi_d$ reduces the enumeration to the eight branch candidates $\{\mathrm{AIK}(\mathbf{x}_d,\mathbf{j},\varphi_d)\}_{\mathbf{j}\in\{-1,+1\}^3}$, whereas the free-redundancy mode used hereafter samples $\varphi$ over $[0,2\pi)$; the proposed selection ranks whichever candidate set results.}

The resulting analytical solutions are subsequently used to generate candidate joint configurations, which serve as initial seeds for the hybrid inverse kinematics framework proposed in this work.

\subsection{LM-based Numerical Inverse Kinematics (NIK)}\label{sec:LM}
Let $\mathbf{q}\in\mathbb{R}^7$ denote joint angles with limits $\mathbf{q}_{\min}\le \mathbf{q}\le \mathbf{q}_{\max}$. Given a desired end-effector pose $(\mathbf{p}_d,\mathbf{R}_d)$, define forward kinematics
$
\mathbf{T}(\mathbf{q})=
\begin{bmatrix}
\mathbf{R}(\mathbf{q}) & \mathbf{p}(\mathbf{q})\\
\mathbf{0}^\top & 1
\end{bmatrix}.
$
We define a 6D pose error using a Lie-algebra orientation residual:
\begin{equation}
\mathbf{e}(\mathbf{q}) \;=\;
\begin{bmatrix}
\mathbf{p}(\mathbf{q})-\mathbf{p}_d\\
\mathrm{Log}\!\left(\mathbf{R}_d^\top \mathbf{R}(\mathbf{q})\right)
\end{bmatrix}
\in\mathbb{R}^6,
\label{eq:pose_error}
\end{equation}
and a weighted least-squares objective
\begin{equation}
\begin{aligned}
&\min_{\mathbf{q}\in\mathcal{Q}}\;
\frac{1}{2}\left\|\mathbf{W}\,\mathbf{e}(\mathbf{q})\right\|_2^2, \quad \mathbf{W} =\mathrm{diag}(\sqrt{w_p}\mathbf{I}_3,\sqrt{w_R}\mathbf{I}_3),
\end{aligned}
\label{eq:ik_objective}
\end{equation}
{where $w_p$ and $w_R$ are the weights of position and rotation, respectively.} The LM update at iterate $\mathbf{q}_t$ solves the damped normal equations:
\begin{equation}
\begin{aligned}
&\left(\mathbf{J}_t^\top \mathbf{W}^\top\mathbf{W}\mathbf{J}_t + \lambda_t \mathbf{I}\right)\Delta\mathbf{q}_t
=
-\mathbf{J}_t^\top \mathbf{W}^\top\mathbf{W}\mathbf{e}(\mathbf{q}_t),
\\
&\mathbf{q}_{t+1}=\Pi_{[\mathbf{q}_{\min},\mathbf{q}_{\max}]}\!\left(\mathbf{q}_t+\Delta\mathbf{q}_t\right),
\end{aligned}
\label{eq:lm_update}
\end{equation}
where $\mathbf{J}_t=\partial\mathbf{e}/\partial\mathbf{q}|_{\mathbf{q}_t}$ and $\Pi$ denotes projection to joint limits.

\subsection{Heuristic-based Candidate Selection}

For a given end-effector pose, AIK typically yields an infinite number of joint solutions due to the presence of multiple kinematic branches and system redundancy. Consequently, a heuristic-based strategy is adopted to select a suitable candidate solution from the AIK solution set. Since these analytical solutions are obtained using a simplified kinematic model, the chosen candidate is subsequently refined by a numerical inverse kinematics solver based on the complete kinematic model.

\subsubsection{Distance-based Criterion}
The first heuristic is based on minimizing the distance in joint space between the candidate solution and the current joint configuration $\mathbf{q}_{\mathrm{init}}$. This criterion is formulated as
\begin{align}
\min_{\substack{
\mathbf{j} \in \{-1,1\}^3 \\
\varphi \in [0,2\pi)
}}
\quad
d = \left\| \mathbf{q}_{\mathrm{init}} - \mathbf{q}(\mathbf{j}, \varphi) \right\|_2^2 .\label{eq:dist_cri}
\end{align}
This criterion favors small joint displacement between consecutive configurations.

\subsubsection{Manipulability-based Criterion}
A manipulability-based criterion \cite{yoshikawa1990translational} penalizes configurations close to kinematic singularities.
\begin{align}
\min_{\substack{
\mathbf{j} \in \{-1,1\}^3 \\
\varphi \in [0,2\pi)
}}
\,
m = -\log \!\left(
\sqrt{
\det\!\left(
\mathbf{J}(\mathbf{q}(\mathbf{j}, \varphi))
\mathbf{J}(\mathbf{q}(\mathbf{j}, \varphi))^\top
\right)
}
\right),\label{eq:manip_cri}
\end{align}
where $\mathbf{J}(\cdot)$ denotes the manipulator Jacobian. 
The logarithmic form improves numerical stability.

\subsubsection{Weighted Criterion}
The two criteria \eqref{eq:dist_cri} and \eqref{eq:manip_cri} are combined into a weighted cost:
\begin{align}
\min_{\substack{
\mathbf{j} \in \{-1,1\}^3 \\
\varphi \in [0,2\pi)
}}
\quad
w_d \,\big( 1 - \mathrm{e}^{-d/\tau_d} \big)
+
w_m \,\big( 1 - \mathrm{e}^{-m/\tau_m} \big), \label{eq:weighted}
\end{align}
where $w_d, w_m$ weight the two terms and $\tau_d, \tau_m$ normalize them to comparable ranges.

\subsection{Conformal-Seeded Hybrid Inverse Kinematics (CS-HIK)}

Even when analytical inverse kinematics (AIK) produces multiple feasible candidates, commonly used {above-mentioned heuristic selection rules} may lead to suboptimal {seed choices for subsequent numerical refinement}. {Although these approaches are effective for coarse candidate filtering, they do not explicitly account for the discrepancy between the ideal analytical model and the calibrated numerical robot model.} In particular, two candidate seeds may exhibit similar heuristic scores but result in drastically different convergence behavior during numerical refinement due to factors such as {parameter offsets, calibration updates,} local nonconvexity, proximity to kinematic singularities, and interactions with joint limits.

To address this limitation, we formulate seed selection as a \emph{risk-controlled prediction} problem. Among the AIK candidates, the objective is to identify the seed that is most likely to achieve successful numerical refinement within a bounded iteration budget. To this end, we introduce a \emph{conformal-calibrated seed selection} module that ranks AIK candidates using a learned predictor of post-Levenberg--Marquardt (LM) refinement difficulty, combined with split conformal prediction. {\color{black}The learned predictor determines the ranking; the conformal layer turns its prediction into a calibrated, distribution-free upper bound that is used as the selection score. Both steps satisfy strict real-time computational constraints.} {The selection mechanism itself is generator- and robot-agnostic: the features \eqref{eq:features} scale with the DoF ($D = 2\,\mathrm{DOF} + 4$), and only a finite seed set and the forward kinematics of the calibrated model are required, so extension to other redundant manipulators reduces to supplying a suitable candidate generator; the enumeration of Section~III-A instantiates this for the offset 7-DoF humanoid class considered here.}

\subsubsection{Hybrid IK with Candidate Enumeration}
AIK produces a finite candidate set $\mathcal{S}=\{\mathbf{q}^{(k)}_0\}_{k=1}^K$ by enumerating discrete kinematic branches and sampling the redundancy parameter (e.g., elbow swivel angle). Each candidate is checked against joint limits. The classical hybrid IK chooses a single seed by a heuristic and then applies LM:
\begin{equation}
\mathbf{q}^{(k)}_* \leftarrow \mathrm{LM}\!\left(\mathbf{q}^{(k)}_0; \mathbf{p}_d,\mathbf{R}_d\right).
\label{eq:lm_refine}
\end{equation}
The overall success depends strongly on selecting a seed that leads to rapid convergence and avoids singularities or joint-limit trapping.

\subsubsection{Conformal-Calibrated Seed Selection}

We introduce a seed-scoring function that predicts the \emph{post-LM normalized residual} from cheap-to-compute seed features.
For each AIK candidate $\mathbf{q}^{(k)}_0$, define a feature vector {$\bar{\mathbf{x}}^{(k)}$ as follows}
\begin{equation}
\bar{\mathbf{x}}^{(k)} \;=\; \phi\!\left(\mathbf{q}^{(k)}_0,\mathbf{q}_{\mathrm{init}}, \mathbf{p}_d,\mathbf{R}_d\right)\in\mathbb{R}^D,
\label{eq:features}
\end{equation}
including initial pose error norms, distance to joint limits, joint-space distance to $\mathbf{q}_{\mathrm{init}}$, and a Jacobian dexterity proxy (e.g., log-determinant of $\mathbf{J}\mathbf{J}^\top$). {Here, $\phi(\cdot)$ is a feature-extraction function that maps each AIK candidate seed and the task information to the feature vector $\bar{\mathbf{x}}^{(k)} \in \mathbb{R}^{D}$.} Let $y^{(k)}$ denote the \emph{true} post-LM normalized error obtained offline:
{
\small
\begin{align}
y^{(k)} =
\max\left(\frac{\|\mathbf{p}(\mathbf{q}^{(k)}_*)-\mathbf{p}_d\|_2}{\varepsilon_p},
           \frac{\|\mathrm{Log}(\mathbf{R}_d^\top\mathbf{R}(\mathbf{q}^{(k)}_*))\|_2}{\varepsilon_R}\right), \label{eq:normalized_label}
\end{align}
}
success $\Leftrightarrow y^{(k)} < 1$, where $(\varepsilon_p,\varepsilon_R)$ are task tolerances.

{To construct the regression dataset, we randomly generate $N_{\mathrm{train}} + N_{\mathrm{calib}}$ IK samples and compute the corresponding feature/label pairs. A linear ridge regressor $\hat{y}(\mathbf{x})$ is first trained offline using the training set $\mathcal{D}_{\mathrm{train}}=\{(\mathbf{x}_i,y_i)\}_{i=1}^{N_{\mathrm{train}}}$. Here, $\mathbf{x}_i$ represents the extracted seed feature vector, while $y_i$ denotes the corresponding post-LM normalized residual defined in} \eqref{eq:normalized_label}{. Subsequently, we consider a calibration set $\mathcal{D}_{\mathrm{cal}}=\{(\mathbf{x}_i,y_i)\}_{i=1}^{N_{\mathrm{calib}}}$ and define conformal residuals $r_i$ using \emph{split conformal prediction} as follows}
\begin{equation}
r_i \;=\; y_i - \hat{y}(\mathbf{x}_i),
\label{eq:conformal_residual}
\end{equation}
and the $(1-\alpha)$ quantile
\begin{equation}
\hat{q}_{\alpha} \;=\; \mathrm{Quantile}_{1-\alpha}\big(\{r_i\}_{i=1}^n\big).
\label{eq:qhat}
\end{equation}
Then for a new seed with features $\bar{\mathbf{x}}^{(k)}$, we obtain an upper confidence bound
\begin{equation}
U^{(k)} \;=\; {\color{black}\max\Big(0,\;} \hat{y}(\bar{\mathbf{x}}^{(k)}) + \hat{q}_{\alpha} {\color{black}\Big)},
\label{eq:ucb}
\end{equation}
{\color{black}the outer floor enforcing $U^{(k)}\ge 0$, consistent with $U^{(k)}$ representing a predicted, necessarily non-negative post-LM error; since the floor can only raise $U^{(k)}$, the coverage statement below is unaffected.} {\color{black}$U^{(k)}$} satisfies marginal coverage under exchangeability assumptions:
\begin{equation}
\mathbb{P}\!\left(y^{(k)} \le U^{(k)}\right)\;\ge\;1-\alpha.
\label{eq:coverage}
\end{equation}
We select the seed that minimizes this conformal upper bound:
\begin{equation}
k^* \;=\; \arg\min_{k\in\{1,\dots,K\}} U^{(k)}.
\label{eq:seed_select}
\end{equation}
This selection is computationally negligible for small $D$ and $K$, enabling real-time deployment. Optionally, we keep the top-$M$ seeds by $U^{(k)}$ as a fail-safe shortlist. {The selection in} \eqref{eq:seed_select} {is a data-dependent operation. Although each $U^{(k)}$ marginally satisfies \eqref{eq:coverage} for a pre-specified index $k$, choosing the minimizing index $k^{*}$ breaks exchangeability between the calibration set and the selected candidate, so \eqref{eq:coverage} does not transfer to the selected seed (a ``winner's curse'' effect). Accordingly, $U^{(k^{*})}$ is not interpreted as a calibrated bound on the selected seed; instead, $U^{(k)}$ is used {\color{black}as a ranking score, and \eqref{eq:coverage} is read as a statement about a pre-specified candidate only}. Selection-conditional coverage would require dedicated selection-conditional or simultaneous conformal procedures, which we leave to future work.}

{\color{black}The role of $\hat q_\alpha$ in the ranking is stated in the following. For candidates not clipped by the floor, adding the same constant to all predictions cannot change their order, so $\operatorname{argsort}_k U^{(k)}$ coincides with $\operatorname{argsort}_k \hat y(\bar{\mathbf x}^{(k)})$: the ranking is determined by the learned predictor alone. The margin enters only through the floor. Since $\hat y$ is an unconstrained linear predictor, it returns $\hat y<0$ for a sizeable fraction of candidates; without the margin, all of them are clipped to $U^{(k)}=0$ and the choice among them falls back to the enumeration order. With $\hat q_\alpha>0$, the candidates with $\hat y>-\hat q_\alpha$ are lifted above the floor and the ordering by $\hat y$ is restored (Section~IV-B, Table~\ref{tab:ablation}). What the conformal layer adds is calibration: for a pre-specified candidate, $U^{(k)}$ bounds the post-LM error in units of the task tolerance at level $1-\alpha$ by \eqref{eq:coverage}, for any predictor and without distributional assumptions, which motivates its use in the small-data regime targeted here. Note that $\hat q_\alpha$ converges to the $(1-\alpha)$-quantile of the residual distribution of $\hat y$ and does not vanish for an imperfect predictor as the dataset grows.}

\subsubsection{Algorithm Overview}
CS-HIK solves IK in three stages: (i) candidate generation, (ii) {\color{black}learned} seed ranking, and (iii) numerical refinement with a bounded fallback. In Line~2 of Alg. \ref{alg:cs_hik}, we transform the desired pose from the world frame to the solver frame (e.g., the shoulder base frame) to match the analytical IK derivation and ensure consistency between AIK and LM refinement. Then, $\mathrm{AIK.Enumerate}(\cdot)$ enumerates a \emph{finite candidate set} $\mathcal{S}=\{\mathbf{q}^{(k)}_0\}_{k=1}^K$. 
For each seed $\mathbf{q}^{(k)}_0$, we compute the feature vector $\mathbf{x}^{(k)}=\phi(\cdot)$ of \eqref{eq:features} (Line~6), and the lightweight predictor $\hat{y}(\mathbf{x})$ estimates the post-LM difficulty of that seed. 
To obtain uncertainty-aware and statistically calibrated selection, we form the conformal upper bound
$
U^{(k)}={\color{black}\max\big(0,\;}\hat{y}(\mathbf{x}^{(k)})+\hat{q}_\alpha{\color{black}\big)}
$
(Line~7), where $\hat{q}_\alpha$ is computed offline on a held-out calibration set.
{For a pre-specified candidate index, $U^{(k)}$ provides a marginal upper bound on post-LM difficulty under standard conformal assumptions (\eqref{eq:coverage}); at deployment, however, we use $U^{(k)}$ only to \emph{rank} candidates rather than to claim coverage for the selected seed (see the remark after \eqref{eq:seed_select}). 
Thus, ranking seeds by increasing $U^{(k)}$ (Line~9) prioritizes seeds predicted to be easiest to refine, which is more robust than greedy heuristic rules in \eqref{eq:dist_cri} and \eqref{eq:manip_cri}. The complete conformal-seeded hybrid IK procedure is summarized in Algorithm~\ref{alg:cs_hik}. }

{
\begin{rem}\label{rem:scope}
Algorithm~\ref{alg:cs_hik} is a \emph{stateless per-query global solver}: each invocation enumerates and selects independently of previous solutions, the intended behavior for point-to-point queries, cold starts, goal-configuration generation, and recovery after target jumps (the settings of Section~IV-B). For continuous tracking, joint-space continuity is obtained either by seeding \eqref{eq:lm_refine} with the previous solution $\mathbf{q}_{t-1}$, supported by construction since the LM stage accepts an arbitrary seed and preserves continuity within its convergence basin, or by delegating smoothness to the planning layer (Section~IV-C-4). Section~IV-C-3 deliberately invokes the stateless mode at every cycle to expose the re-selection behavior.
\end{rem}}

\begin{algorithm}[t]
\caption{Conformal-Seeded Hybrid Inverse Kinematics (CS-HIK)}
\label{alg:cs_hik}
\begin{algorithmic}[1]
\REQUIRE Desired end-effector pose $(\mathbf{p}_d,\mathbf{R}_d)$, joint limits $[\mathbf{q}_{\min},\mathbf{q}_{\max}]$, current state $\mathbf{q}_{\mathrm{init}}$, conformal model $(\hat{y},\hat{q}_{\alpha})$.
\ENSURE Optimized configuration $\mathbf{q}^*$ (or failure)
\STATE Compute transform in solver frame:  $\mathbf{T}_{07,d} \leftarrow \mathbf{T}_{0w}^{-1} \mathbf{T}_{7w,d}$
\STATE \textbf{\textit{Step 1: Candidate generation (AIK enumeration)}}
\STATE $\mathcal{S} \leftarrow \{\mathbf{q}^{(k)}_0\}_{k=1}^K \gets \mathrm{AIK.Enumerate}(^0\mathbf{T}_{7,d}, \mathbf{q}_{\min}, \mathbf{q}_{\max})$
\STATE \textbf{\textit{Step 2: Conformal seed scoring}}
\FOR{$k=1$ to $K$}
\STATE \quad $\bar{\mathbf{x}}^{(k)} \leftarrow \phi(\mathbf{q}^{(k)}_0,\mathbf{q}_{\mathrm{init}},\mathbf{p}_d,\mathbf{R}_d)$
\STATE \quad $U^{(k)} \leftarrow {\color{black}\max\big(0,\;}\hat{y}(\bar{\mathbf{x}}^{(k)}) + \hat{q}_{\alpha}{\color{black}\big)}$
\ENDFOR
\STATE \quad $\pi \leftarrow \mathrm{argsort}_k(U^{(k)})$ \COMMENT{rank seeds by conformal {score}}
\STATE \textbf{\textit{Step 3: Numerical refinement (LM) with fail-safe}}
\FOR{$m=1$ to $\min(M,K)$}
\STATE \quad $\mathbf{q}^* \leftarrow \mathrm{LM}\!\left(\mathbf{q}^{(\pi_m)}_0;\mathbf{p}_d,\mathbf{R}_d\right)$
\IF{$\mathbf{q}^*$ meets tolerances and limits}
\STATE \quad \textbf{return} $\mathbf{q}^*$
\ENDIF
\ENDFOR
\STATE \textbf{return failure}
\end{algorithmic}
\end{algorithm}


\begin{table}[]
\centering
\caption{Performance comparison of inverse kinematics methods}%
\label{tab:ik_comparison}
\begin{threeparttable}
\footnotesize
\renewcommand{\arraystretch}{1.05}
\begin{tabular}{lllcccc}
\toprule
\multicolumn{3}{l}{\cellcolor[HTML]{EFEFEF}{Methods}}                                                                              &\cellcolor[HTML]{EFEFEF} \begin{tabular}[c]{@{}c@{}}{Success}\\{rate}\\ (\%)\end{tabular}     & \cellcolor[HTML]{EFEFEF}\begin{tabular}[c]{@{}c@{}}{Exec.}\\{time}\\ ($\mu \mathrm{s}$)\end{tabular} & \cellcolor[HTML]{EFEFEF}\begin{tabular}[c]{@{}c@{}} {Pos.}\\{error}\\ {($\mu\mathrm{m}$)}\end{tabular}   & \cellcolor[HTML]{EFEFEF}\begin{tabular}[c]{@{}c@{}} {Rot.}\\{error}\\ {($\mu\mathrm{rad}$)}\end{tabular} \\ \midrule
\multicolumn{3}{l}{{AIK} ~\eqref{eq:aik}}                                                                                 & 100 & 0.227                      & 9640                          & 2520                             \\ 
\multicolumn{3}{l}{{NIK} ~\eqref{eq:lm_update}}                                                                                 & 65.37                         & 283.8                         & 285                           & 489                              \\
\multicolumn{3}{l}{{KDL-LMA}\cite{bruyninckx2001open}}                                                                                 & \iffalse99.36  - \fi12.00 \tnote{a}                      & 154.1                         & 2.8                           & 1649       \\
\multicolumn{3}{l}{{KDL-NR-JL}\cite{bruyninckx2001open}}                                                                                 & 21.67                       & 780.35                         & 0.03                           & 0.4       \\
\multicolumn{3}{l}{{TracIK} \cite{beeson2015trac}}                                                                              & 99.38                        & 728.70                         & 2.9                    & 4.2    \\
\multicolumn{3}{l}{{PyRoki} \cite{kim2025pyroki}}                                                                              & 80.45                         & 448.0                           & 6586                          & 16.3                 \\
\multicolumn{3}{l}{{QuIK \tnote{c}} \cite{lloyd2022fast}}                                                                              & \iffalse 99.96 - \fi16.89 \tnote{a}                     & 3.53                        & - \tnote{b}                        & -     \tnote{b}
\\
\midrule
                                &\cellcolor[HTML]{EFEFEF}$\Delta\varphi$                                          &\cellcolor[HTML]{EFEFEF}         & \cellcolor[HTML]{EFEFEF}                     &  \cellcolor[HTML]{EFEFEF}               &    \cellcolor[HTML]{EFEFEF}                 & \cellcolor[HTML]{EFEFEF}                           \\
                                &                                 &         dist. \eqref{eq:dist_cri}          & 96.41                         & 52.13                         & 3.2                           & 7.4                              \\
                                &                                          & {manip. \eqref{eq:manip_cri}}          & 96.61                         & 50.83                         & 3.0                           & 7.5                              \\
                                &                                          & weighted \eqref{eq:weighted}                   & 96.41                         & 53.40                         & 3.2                           & 7.4                              \\
                                &                                          & {rank-1 CS} \eqref{eq:seed_select}                    & 97.02                         & 13.13                         & 9.4                           & 9.9                              \\
                                & \multirow{-5}{*}{{$\dfrac{\pi}2$}} & \cellcolor[HTML]{EFEFEF}CS \eqref{eq:seed_select}& \cellcolor[HTML]{EFEFEF}97.26 & \cellcolor[HTML]{EFEFEF}44.38 & \cellcolor[HTML]{EFEFEF}2.5   & \cellcolor[HTML]{EFEFEF}6.4      \\ \cline{2-7} 
                                &                                      &    dist. \eqref{eq:dist_cri}          & 99.19                         & 32.66                         & 3.7                           & 5.5                              \\
                                &                                          & {manip. \eqref{eq:manip_cri}}          & 99.67                         & 32.62                         & 2.6                           & 4.9                              \\
                                &                                          & weighted \eqref{eq:weighted}                   & 99.19                         & 34.79                         & 3.6                           & 5.6                              \\
                                &                                          & {rank-1 CS} \eqref{eq:seed_select}                    & 98.90                        & 18.00                         & 9.6                           & 10.7                              \\
                                & \multirow{-5}{*}{{$\dfrac{\pi}3$ }} & \cellcolor[HTML]{EFEFEF}CS \eqref{eq:seed_select}& \cellcolor[HTML]{EFEFEF}99.02 & \cellcolor[HTML]{EFEFEF}27.86 & \cellcolor[HTML]{EFEFEF}2.6   & \cellcolor[HTML]{EFEFEF}3.1      \\ \cline{2-7} 
                                &                                        &  dist. \eqref{eq:dist_cri}          & 99.49                         & 34.44                         & 4.7                           & 7.0                              \\
                                &                                          & {manip. \eqref{eq:manip_cri}}          & 99.64                         & 32.76                         & 2.8                           & 4.5                              \\
                                &                                          & weighted \eqref{eq:weighted}                   & 99.43                         & 37.00                         & 4.1                           & 6.4                              \\
                                &                                          & {rank-1 CS} \eqref{eq:seed_select}                   & 99.93                         & 31.65                         & 9.1                           & 10.1                              \\
                                & \multirow{-5}{*}{{$\dfrac{\pi}6$}}  & \cellcolor[HTML]{EFEFEF}CS \eqref{eq:seed_select}& \cellcolor[HTML]{EFEFEF}99.95 & \cellcolor[HTML]{EFEFEF}27.35 & \cellcolor[HTML]{EFEFEF}3.2  & \cellcolor[HTML]{EFEFEF}1.7     \\ \cline{2-7} 
                                &                                         & dist. \eqref{eq:dist_cri}          & 99.42                         & 49.42                         & 5.2                           & 9.0                              \\
                                &                                          & {manip. \eqref{eq:manip_cri}}          & 99.67                         & 25.02                         & 2.8                           & 5.2                              \\
                                &                                          & weighted \eqref{eq:weighted}                   & 99.37                         & 52.13                         & 4.5                           & 8.0                              \\
                                &                                          & {rank-1 CS} \eqref{eq:seed_select}                  & 99.97                         & 33.21                         & 9.3                           & 10.1                              \\
                                & \multirow{-5}{*}{{$\dfrac{\pi}9$}}  & \cellcolor[HTML]{EFEFEF}CS \eqref{eq:seed_select} & \cellcolor[HTML]{EFEFEF}{\textbf{100}} & \cellcolor[HTML]{EFEFEF}{\textbf{39.52}} & \cellcolor[HTML]{EFEFEF}3.4  & \cellcolor[HTML]{EFEFEF}1.9     \\ \cline{2-7} 
                                &                                        &  dist. \eqref{eq:dist_cri}          & 99.19                         & 57.61                         & 5.6                           & 10.8                             \\
                                &                                          & {manip. \eqref{eq:manip_cri}}          & 99                            & 55.12                         & 2.8                           & 6.2                              \\
                                &                                          & weighted \eqref{eq:weighted}                   & 99.19                         & 65.00                         & 4.8                           & 9.6                              \\
                                &                                          & {rank-1 CS}  \eqref{eq:seed_select}                  & 99.97                         & 42.45                         & 9.1                         & 10.0                              \\
\multirow{-20}{*}{\rotatebox{90}{{HIK (Ours) {\texttt{@} $M = 4$}}}} & \multirow{-4}{*}{{$\dfrac{\pi}{12}$}} & \cellcolor[HTML]{EFEFEF}CS  \eqref{eq:seed_select}& \cellcolor[HTML]{EFEFEF}100   & \cellcolor[HTML]{EFEFEF}49.12 & \cellcolor[HTML]{EFEFEF}3.4  & \cellcolor[HTML]{EFEFEF}1.8     \\ \toprule
\end{tabular}
\begin{tablenotes}
\footnotesize
\item[a] Primarily caused by joint limit violations.
\item[b] QuIK supports only the Denavit--Hartenberg (DH) convention; hence, the reported position and orientation errors strongly depend on the discrepancies between the URDF and DH representations.
\item[c] DH-table support only.
\end{tablenotes}

\end{threeparttable}
\end{table}

\section{Experimental results}
\subsection{{Simulation and experiment setup}}
{This section evaluates the proposed IK framework in terms of robustness, computational efficiency, and solution accuracy using Monte-Carlo simulations over a densely sampled workspace. Experiments on a humanoid arm demonstrate real-time feasibility under practical conditions.}

{Regarding the conformal-seeded mechanism, the feature vector consists of the initial end-effector position and orientation error norms (2 features), distances to the joint limits for each joint ($2\times\mathrm{DOF}$ features), joint-space distance to $\mathbf{q}_{\mathrm{init}}$ (1 feature), and a Jacobian dexterity metric (1 feature). For the 7-DOF manipulator considered in this study, the total feature dimension is therefore $D = 18$. The feature vectors are standardized using the mean and standard deviation computed from the training dataset before being used for regression and conformal calibration. The regression target is defined in \eqref{eq:normalized_label}.}

{A lightweight Ridge regression model is employed as the conformity predictor due to its low computational overhead and stable inference performance for real-time applications. The training and calibration datasets contain 80 and 20 samples, respectively. The Ridge regularization coefficient is set to $\lambda = 0.01$, while the conformal confidence level is selected as $\alpha = 0.05${\color{black}, which yields $\hat q_\alpha = 0.119$ on the calibration set}. 
{\color{black}The position and orientation tolerances that define the success label \eqref{eq:normalized_label} are $\varepsilon_p = 10^{-3}\,\mathrm{m}$ and $\varepsilon_R = 10^{-3}\,\mathrm{rad}$, respectively.}
For the LM refinement stage, the weighting coefficients for the position and orientation errors are chosen as $w_p = 10$ and $w_R = 1$, respectively, while the maximum number of LM iterations is limited to 300{\color{black}, and LM is stopped once the cost function in \eqref{eq:ik_objective} drops below $10^{-8}$}. The heuristic weights are set to $w_d = 0.4$, $w_m = 0.6$, and the normalization scales to $\tau_d = 8.237$, $\tau_m = 4.419$. The complete offline construction of the conformity predictor, which automatically generates the dataset (workspace sampling and label generation via LM refinement) and then performs the Ridge fitting and conformal calibration, requires approximately \SI{3}{\second} on a standard desktop CPU, of which the Ridge fit itself takes only several milliseconds; the resulting predictor is global and reused across the entire workspace.}

\subsection{Monte-Carlo simulations}
\begin{table}[]
    \centering
    \footnotesize
    \caption{{\color{black}Ablation of the ranking score. All rows use the identical predictor $\hat y$ (same training samples and features) and differ only in the score sorted in Line~9 of Algorithm~\ref{alg:cs_hik}.}}
    \renewcommand{\arraystretch}{1.05}
    \setlength{\tabcolsep}{4pt}
    \begin{tabular}{l c c c}
    \hline
    \cellcolor[HTML]{EFEFEF}{Ranking score} & \cellcolor[HTML]{EFEFEF}$\Delta\varphi$ & \cellcolor[HTML]{EFEFEF}\begin{tabular}[c]{@{}c@{}}{Success rate (\%)}\\{\texttt{@} $M = 1$} \end{tabular}  &  \cellcolor[HTML]{EFEFEF}\begin{tabular}[c]{@{}c@{}}{Success rate (\%)}\\{\texttt{@} $M = 3$} \end{tabular}   \\
    \hline
    {\color{black}Raw $\hat{y}$}      & \color{black}$\pi/6$ & \color{black}99.93 & \color{black}99.95 \\
    {\color{black}Floored $\max(0,\hat{y})$} & $\pi/6$ & {99.38} & {99.94} \\
    \cellcolor[HTML]{EFEFEF}CS-HIK, {\color{black}$\max(0,\hat{y}+\hat{q}_\alpha)$}         & \cellcolor[HTML]{EFEFEF}$\pi/6$ & \cellcolor[HTML]{EFEFEF}{99.93} & \cellcolor[HTML]{EFEFEF}{99.95} \\
    \hline
    {\color{black}Raw $\hat{y}$}      & $\color{black}\pi/9$ & \color{black}99.97 & \color{black}100 \\
    {\color{black}Floored $\max(0,\hat{y})$} & $\pi/9$ & {99.26} & {99.99} \\
    \cellcolor[HTML]{EFEFEF}CS-HIK, {\color{black}$\max(0,\hat{y}+\hat{q}_\alpha)$}         &\cellcolor[HTML]{EFEFEF} $\pi/9$ & \cellcolor[HTML]{EFEFEF}{99.97} & \cellcolor[HTML]{EFEFEF}{100} \\
    \hline
    \end{tabular}
    \label{tab:ablation}
\end{table}

\begin{figure}
    \centering
    \includegraphics[width=1\linewidth]{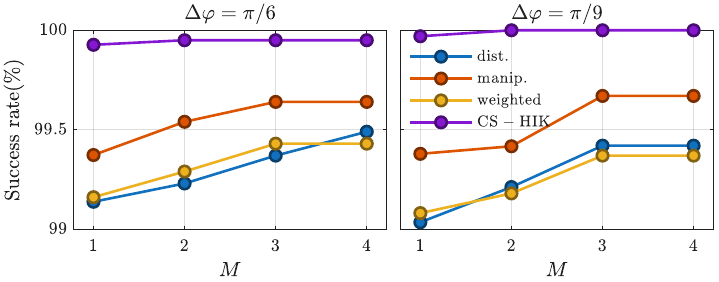}
    \caption{{Success rate comparison under different swivel-angle sampling intervals $\Delta\varphi$, varying numbers of candidate seeds, and varying used methods.}}
    \label{fig:success_rate_diff}
\end{figure}

In the Monte-Carlo simulation, the robot workspace is uniformly sampled by discretizing the joint space
using linear spacing for each joint. This results in a total of
$ 2.5 \times 10^6$
test configurations.
For each sampled joint configuration, the corresponding end-effector pose is computed via forward kinematics
and subsequently used as the target input for inverse kinematics evaluation.
{To ensure a fair evaluation, the ground-truth joint configuration is not provided to the solver; instead, the initial configuration is randomly generated within the joint limits.}
Table~\ref{tab:ik_comparison} summarizes the quantitative comparison of the inverse kinematics methods.

\begin{figure*}
    \centering
    \includegraphics[width=\linewidth]{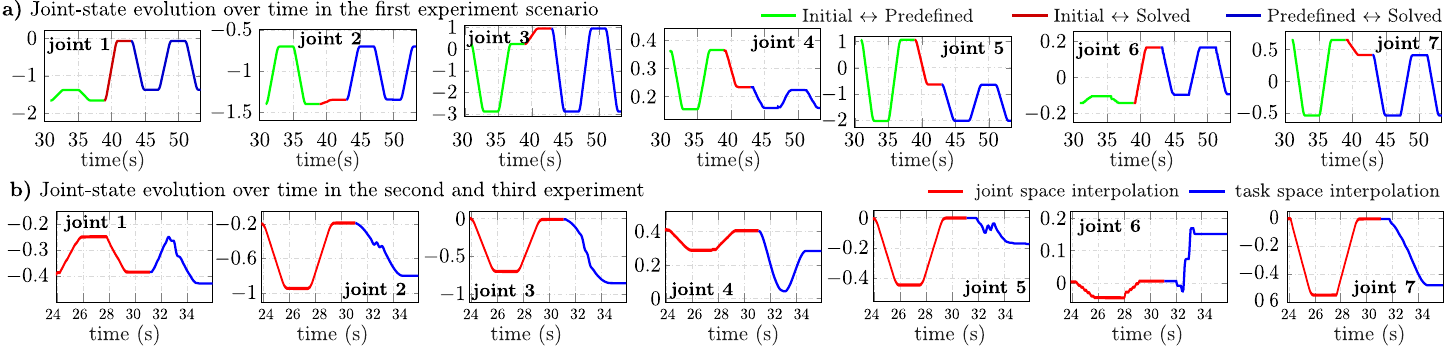}
    \caption{{Joint-state evolution over time in three experiment scenarios.}}
    \label{fig:total}
\end{figure*}

{
AIK achieves a $100\%$ success rate at sub-$\mu\mathrm{s}$ execution time, but the DH--URDF modeling discrepancy leaves large pose errors; NIK is far more accurate yet less robust and slower.
For HIK methods, decreasing the swivel-angle sampling step $\Delta\varphi$ generally improves the success rate by generating better analytical seeds for numerical refinement. Finer sampling raises the AIK enumeration cost but reduces the LM iteration burden, a trade-off between the two stages. For heuristic seed-selection methods, including distance-based, manipulability-based~\eqref{eq:manip_cri}, and weighted strategies, candidate solutions are selected without considering the convergence behavior of the subsequent Levenberg--Marquardt (LM) refinement. As a result, the success rate becomes inconsistent and does not necessarily improve with smaller $\Delta\varphi$; for example, the manipulability criterion drops from $99.67\%$ at $\Delta\varphi = \pi/9$ to $99\%$ at $\pi/12$, while the distance criterion decreases from $99.42\%$ to $99.19\%$. In contrast, the proposed CS-HIK explicitly predicts post-LM convergence behavior during seed selection, as described in \eqref{eq:seed_select}. By accounting for the numerical solver characteristics during seed selection, CS-HIK improves monotonically and reaches $100\%$ success at $\Delta\varphi = \pi/9$ and $\pi/12$ while maintaining real-time performance with an execution time of $39.52\,\mu\mathrm{s}$.} We further compare the proposed approach with several widely used IK solvers, including KDL-LMA, KDL-NR-JL~\cite{bruyninckx2001open}, TracIK~\cite{beeson2015trac}, PyRoki~\cite{kim2025pyroki}, and QuIK~\cite{lloyd2022fast}. As summarized in Table~\ref{tab:ik_comparison}, the proposed method achieves higher success rates with significantly lower execution time.

{Fig.}~\ref{fig:success_rate_diff} { illustrates the effect of the conformal-based seed selection mechanism on the LM refinement process. Although the heuristic-based approaches achieve relatively high success rate due to the quality of the quasi-AIK candidates, they still require larger fail-safe budgets $M$ to reach performance saturation. In contrast, the proposed conformal-seeded mechanism achieves higher success rates using only the first-ranked candidate and typically requires only $M=2$ to reach the saturation region. These results indicate that {\color{black}the learned difficulty ranking} more effectively prioritizes reliable initial seeds for numerical refinement.}
{\color{black}Table~\ref{tab:ablation} isolates the ranking score. All rows use the identical ridge predictor $\hat y$, fit once on the same $N_{\mathrm{train}}=80$ samples and features; the $N_{\mathrm{calib}}=20$ calibration samples only provide $\hat q_\alpha$ for the CS-HIK rows and never refit $\hat y$. The rows differ only in the score sorted in Line~9 of Algorithm~\ref{alg:cs_hik}: the raw prediction $\hat y$, the floored prediction $\max(0,\hat y)$, or the conformal bound \eqref{eq:ucb}. Ranking by the raw prediction {gives the same success rates as CS-HIK at every $\Delta\varphi$ and $M$}: the constant $\hat q_\alpha$ does not reorder untied candidates, so the ranking quality comes from the learned predictor, not from the conformal margin. The floored rows are lower because 22.2\% of all candidates receive $\hat y<0$ and are clipped to $U^{(k)}=0$ without the margin; in 41.0\% of the queries the first rank is then a tie broken by enumeration order. With $\hat q_\alpha>0$, only 0.03\% of the candidates remain clipped and the ordering by $\hat y$ is restored. The conformal layer thus adds calibration at negligible cost, while the gains over the heuristic rules in Table~\ref{tab:ik_comparison} and Fig.~\ref{fig:success_rate_diff} are due to the learned difficulty predictor.}
{\color{black}The property the conformal layer does claim, the bound \eqref{eq:coverage}, can be checked on the same set. For a pre-specified candidate (the first enumerated candidate of each query), the empirical coverage of \eqref{eq:coverage} is 98.2 at $1-\alpha=95\%$. For the selected candidate $k^*$, for which \eqref{eq:coverage} is not guaranteed, the empirical coverage is 99.6\%; $U^{(k^*)}<1$ holds for 100\% of the queries, of which 99.8\% succeed. The bound is thus informative in practice and costs one addition per candidate.}

\subsection{Experimental implementation}

\begin{figure}
    \centering
    \includegraphics[width=\linewidth]{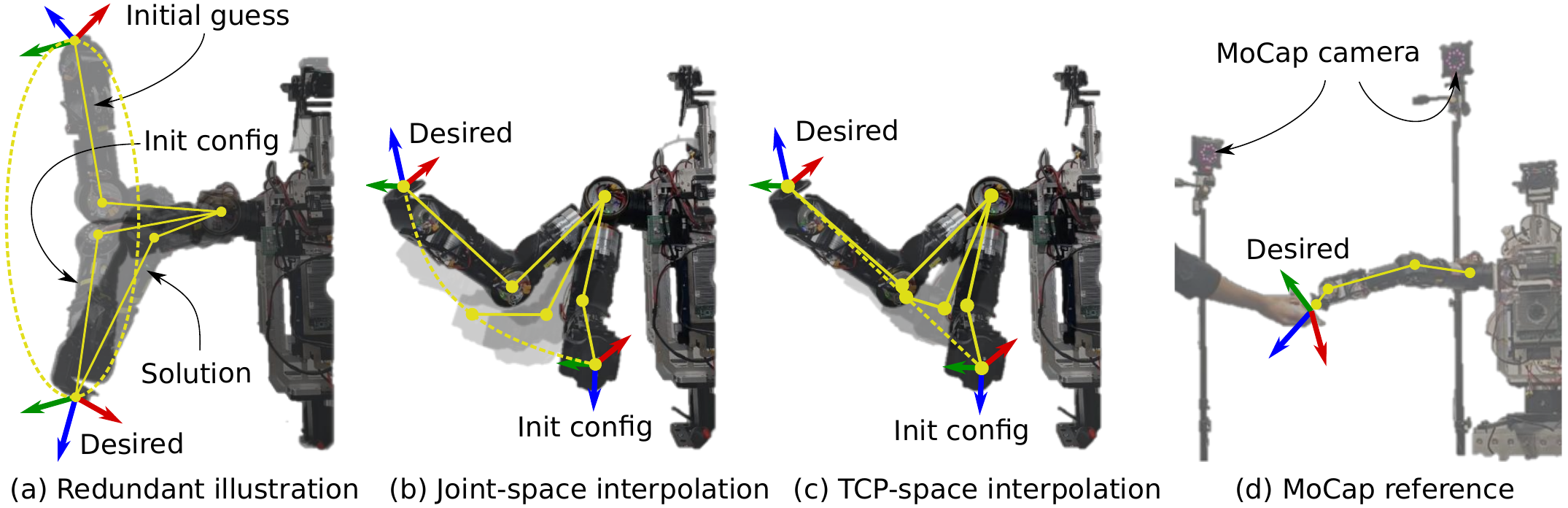}
    \caption{Experimental validation of the proposed (CS-HIK) on a humanoid robot arm. 
    (a) {Redundant} illustration showing that multiple joint configurations can achieve the same desired end-effector pose. 
    (b) Joint-space interpolation between an initial configuration and the IK solution. 
    (c) TCP-space interpolation generating a Cartesian straight-line trajectory. 
    (d) Motion capture (MoCap)-based reference for end-effector tracking evaluation. The experiments are shown at https://youtu.be/aeiBmw1XRbw.}
    \label{fig:experiment}
\end{figure}

For experimental validation, the proposed hybrid inverse kinematics framework is implemented on a humanoid robot arm. 
The manipulator is driven by harmonic-drive actuators connected through an EtherCAT network, enabling high-bandwidth real-time control. At the low level, the robot is controlled using a computed-torque controller with friction compensation to ensure accurate trajectory tracking \cite{siciliano2009robotics}. 

\subsubsection{{Scenario 1: Redundant configuration demonstration}}

Fig.~\ref{fig:total}(a) and Fig.~\ref{fig:experiment}(a) 
illustrate the redundancy of the humanoid manipulator. 
Starting from an initial configuration, the proposed IK solver computes a different joint configuration that achieves the same desired end-effector pose.

\subsubsection{{Scenario 2:} Joint-space interpolation}

In this scenario, motion is generated using joint-space interpolation. 
Starting from an initial configuration, the robot moves to the IK solution using cubic polynomial interpolation in joint space, as shown in Fig.~\ref{fig:experiment}(a, b). 
This yields smooth joint trajectories (Fig.~\ref{fig:total}) but a curved end-effector path due to the nonlinear kinematics.

\subsubsection{{Scenario 3}: Task-space interpolation}

In this scenario, a Cartesian trajectory is specified for the Tool Center Point (TCP). {The solver is invoked in its stateless per-query mode (Remark~\ref{rem:scope}) at every interpolation step: the full candidate set is re-enumerated and re-selected each cycle, with no information from the previous solution.}
As shown in Fig.~\ref{fig:experiment}(c), the resulting end-effector motion follows a nearly straight-line path in Cartesian space. {Small joint-space jitter appears at isolated instants that coincide with a change of the selected candidate between consecutive cycles (Fig.~\ref{fig:total}(b)): adjacent targets may select a different branch $\mathbf{j}$ or an adjacent swivel sample, producing a finite joint-space step of at most $0.045$\,rad per cycle, while every solution meets the task tolerances at the micrometer level of Table~\ref{tab:ik_comparison}. The jitter is thus a re-selection effect of stateless per-cycle solving among redundant candidates, not convergence failure or reduced accuracy; in deployment, continuity is restored by seeding from the previous solution or by the planning layer (Remark~\ref{rem:scope}), as demonstrated in Scenario~4.}

\subsubsection{{Scenario 4:} MoCap-based tracking}

Finally, a motion capture system is used to generate reference trajectories for end-effector tracking. The motion planning is implemented in joint space using the receding-horizon waypoint MPC of \cite{beck2024model}. 
As shown in Fig.~\ref{fig:experiment}(d), the robot follows the MoCap reference using the proposed CS-HIK framework, demonstrating its capability to handle real world experiments. {This scenario reflects the intended deployment architecture: the solver supplies goal configurations while the planning layer enforces joint-space smoothness, and consequently the tracking motion remains smooth without the re-selection effect discussed in Scenario~3.}

\section{Conclusion}

This paper presents a {fast and robust} hybrid inverse kinematics framework for offset redundant manipulators that {combines quasi-analytical solutions with LM numerical refinement}. {The proposed framework uses a reduced set of quasi-AIK candidate solutions as initial seeds and introduces a {\color{black}learned seed-selection mechanism, a lightweight predictor of post-refinement difficulty wrapped in a conformal calibration layer,} to improve convergence robustness and consistency.} {Extensive Monte-Carlo evaluations and comparisons with representative IK approaches demonstrate improved robustness, accuracy, and computational efficiency.} {Experimental validation on a 7-DoF humanoid arm confirms real-time applicability, and CS-HIK provides an efficient and reliable per-query IK solution for offset redundant arms of the humanoid class.} 
{Future work will integrate explicit continuity handling for the continuous-tracking use case, exploit the exposed arm-angle interface for task-level redundancy objectives such as obstacle avoidance, and validate the generator-agnostic seed selection on additional redundant manipulators.}


\bibliographystyle{IEEEtran}
\bibliography{refs}

\vfill

\end{document}